\documentclass[letterpaper, 10 pt, conference]{ieeeconf}  

\IEEEoverridecommandlockouts                              

\usepackage{amsmath}    
\usepackage[final]{graphicx}
\usepackage[lofdepth,lotdepth]{subfig}
\usepackage{tabularx}
\usepackage[font=footnotesize,labelfont=footnotesize]{caption}
\usepackage{verbatim}   
\usepackage{color}      
\usepackage{hyperref} 
\usepackage{multirow}
\usepackage{multicol}
\usepackage{graphicx}
\usepackage{rotating}
\usepackage{array}
\usepackage{xspace}
\usepackage{indentfirst}
\usepackage{amssymb}
\usepackage{float}
\usepackage{algorithm}
\usepackage[algo2e]{algorithm2e}
\usepackage{algorithmic}
\usepackage{comment}
\usepackage{sidecap}
\usepackage{booktabs}
\usepackage{breqn}
\usepackage{cite}
\usepackage{mathtools}
\usepackage{graphicx}
\graphicspath{{./figures/}}
\usepackage{caption}
\usepackage{float}
\usepackage{lipsum}
\usepackage{xcolor}
\usepackage{siunitx}
\usepackage{xspace}
\usepackage{bm}
\usepackage{gensymb}
\let\labelindent\relax
\usepackage[inline]{enumitem}
\usepackage{paralist}
\usepackage{verbatim}
\usepackage{soul}

\newcommand{\hmmm}[1]{\textcolor{orange}{#1}}

\newcommand\norm[1]{\left\lVert#1\right\rVert}
\DeclareMathOperator*{\argmin}{argmin} 

\newcommand{\gobble}[1]{}

\newcounter{tecounter}
\title{\LARGE \bf
Experimental Evaluation of 3D-LIDAR Camera Extrinsic Calibration 
}

\author{Subodh Mishra$^{1}$, Philip R. Osteen$^{2}$, Gaurav Pandey$^{3}$ and Srikanth Saripalli$^{1}$
\thanks{$^{1}$with the Department of Mechanical Engineering, Texas A\&M University
        {\tt\small subodh514@tamu.edu}}%
\thanks{$^{2}$with the Army Reasearch Lab, USA}%
\thanks{$^{3}$with the Ford Motor Company, USA}%
}

\begin{document}

\maketitle
\thispagestyle{empty}
\pagestyle{empty}


\begin{abstract}In this paper we perform an experimental comparison of three different target based 3D-LIDAR camera calibration algorithms. We briefly elucidate the mathematical background behind each method and provide insights into practical aspects like ease of data collection for all of them. We extensively evaluate these algorithms on a sensor suite which consists multiple cameras and LIDARs by assessing their robustness to random initialization and by using metrics like Mean Line Re-projection Error (MLRE) and Factory Stereo Calibration Error. We also show the effect of noisy sensor on the calibration result from all the algorithms and conclude with a note on which calibration algorithm should be used under what circumstances.
\end{abstract}


\section{Introduction}
3D-LIDARs and cameras are ubiquitous to robots. Cameras provide color, texture and appearance information which LIDARs lack and LIDARs provide depth information which cameras lack. Virtually all modern autonomy stacks use multiple distinct types of sensors to represent and interact in the external environment, and many high-level autonomy behaviors (multi-modal object detection, state-estimation, mobile manipulation, etc.) depend on accurate calibrations between sensors, such that data from all sensors can be expressed in a common spatial frame of reference. Yet, there is still no unified approach for calibrating the various sensors present on most autonomous systems. This has motivated research for estimation of extrinsic calibration between various sensors, such as 3D-LIDARs and cameras. Although this area has seen contributions from various robotics labs and research groups, a comprehensive work which analyzes commonly used methods and provides experimental evaluation is lacking.  In this work, we experimentally evaluate commonly used methods for estimating 3D-LIDAR-to-camera extrinsic calibration, offering insights into the strengths and weaknesses of various formulations and providing interesting avenues for further work.

\begin{figure}[!ht]
    \centering
    \includegraphics[scale=0.25]{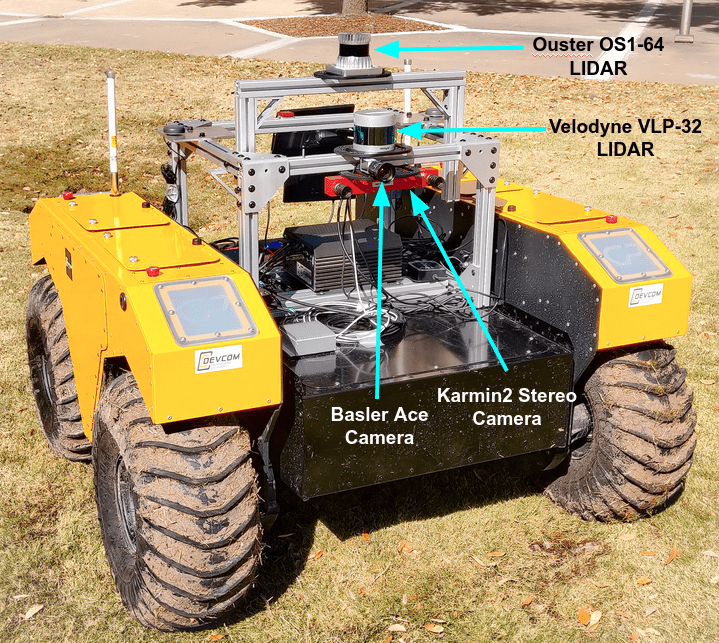}
    \caption{\textbf{Experimental Platform:} Clearpath Robotics Warthog UGV with $a).$ Ouster OS1 LIDAR, $b).$ Velodyne VLP-32 LIDAR, $c).$ Karmin2 Stereo Camera \& $d).$ Basler Ace Camera
    }
    \label{fig:warthogwithsensors}
\end{figure}

\subsection{Literature Survey}
Existing 3D-LIDAR camera extrinsic calibration algorithms can be broadly classified into target based \cite{unnikrishnan2005}, \cite{Huang2009}, \cite{PANDEY2010336}, \cite{Zhou201206}, \cite{Zhou201810}, and targetless approaches \cite{PANDEY201409}, \cite{Levinson201306}, \cite{scaramuzza}, \cite{Taylor201605}.  A target based approach requires a known object in the sensors' common Field of View (FoV) to establish geometric constraints between features detected across the calibrated sensors. While targetless approaches have the obvious advantage of not requiring any special environmental augmentation, reliable accurate data association across modalities is still an open research problem. Although, for many purposes, approximate data association is sufficient (for example, using a camera-based object detection to project a semantic label to a cluster of lidar points), calibration is a problem which uniquely demands metric accuracy above all else. Therefore, approximate data associations with no prior initialization will lead to poor calibration results, which is why even targetless calibration techniques still depend on accurate target based calibration as a precursor. While the ultimate goal of our research is to enable accurate online targetless calibration in arbitrary environments, here we focus on the offline target based calibration to identify the best in class approaches to data association and multi-sensor optimization.



The solution to target based 3D-LIDAR camera extrinsic calibration problem is inspired from target based 2D-LIDAR camera extrinsic calibration \cite{Strauss1995}, \cite{Zhang1389752}, \cite{Ruben7139700}, \cite{Naroditsky2011}, \emph{etc}. The geometric constraint used in \cite{Zhang1389752} is easy to use in 3D-LIDAR camera calibration scenario and has been exploited in the work presented in \cite{unnikrishnan2005} \& \cite{Huang2009} and extended to 3D-LIDAR omnidirectional camera calibration in \cite{PANDEY2010336}.  \cite{Zhou201206} present a 3D-LIDAR camera calibration technique in which the rotation matrix is estimated first and then a \textit{point to plane} constraint (similar to ones in \cite{unnikrishnan2005}, \cite{Huang2009}, \cite{PANDEY2010336}) is used to determine the transformation parameters. \cite{Zhou201810} adds more geometric constraints by introducing line correspondences in addition to the previously used plane correspondences. \cite{unnikrishnan2005}, \cite{Huang2009}, \cite{PANDEY2010336}, \cite{Zhou201206} and \cite{Zhou201810} use a planar target with a checkerboard pattern. \cite{Bonnifait4648067} and \cite{butvelodyne} present calibration methods that use a rigid plane with one and four circular perforations respectively. The methods involve detection of circle in both the image and the 3D-LIDAR data and using the geometric constraints to determine the SE(3) transformation between the 3D-LIDAR and the camera. All of the aforementioned methods which use a single planar target require several observations from geometrically distinct view points. \cite{Geiger} presents a single view calibration technique, but uses several checkerboard planes. In addition to the \textit{point to plane} geometric constraint that form the basis of methods described in \cite{unnikrishnan2005}, \cite{Huang2009} and \cite{PANDEY2010336}, the \textit{}{point to back-projected plane} constraint has been exploited in works described in \cite{Ruben7139700} and \cite{Naroditsky2011}, but only for calibrating 2D-LIDAR camera systems. \cite{subodhIV2020} exploits the \textit{point to back-projected plane} constraint for cross calibrating 3D LIDAR camera pair. The methods described so far are pair-wise 3D-LIDAR camera calibration techniques. For robots with multiple cameras and LIDARs, joint calibration techniques like \cite{jointCalNg} and \cite{Owens_msg-cal:multi-sensor} have been found to be useful. Most target based 2D/3D-LIDAR camera extrinsic calibration methods, which use one or more planar surfaces, use checkerboards or ArUco \cite{articleAruco1} or AprilTags \cite{olson2011tags} for easy detection of planar target in the camera. In cases where such markers are not used, perforated \cite{Bonnifait4648067} \& \cite{butvelodyne}  or spherical targets \cite{spherical2018} are utilized. 

\subsection{Contributions}
\begin{figure}[!ht]
\centering
    \includegraphics[scale=0.08]{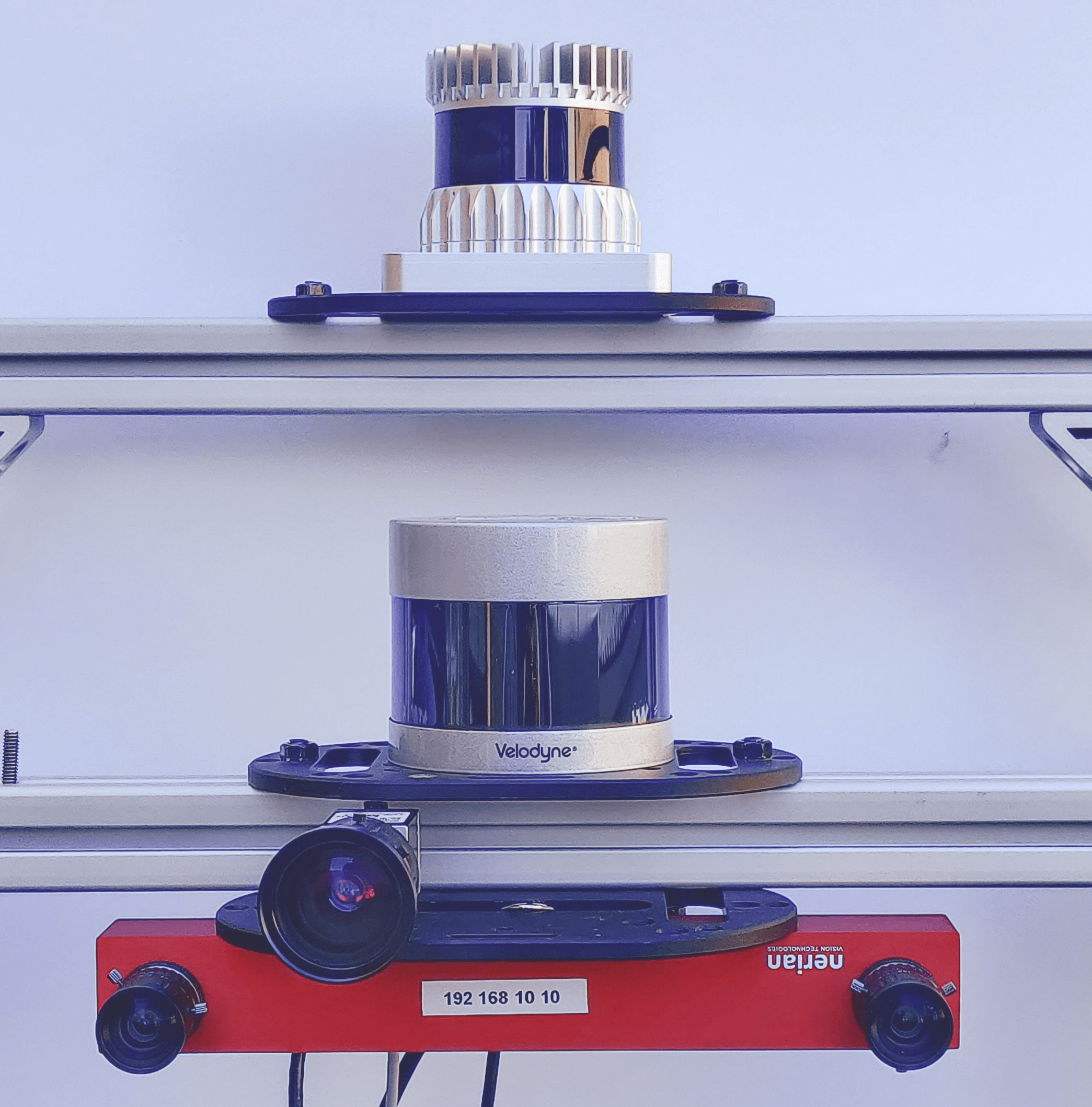}
    \caption{\textbf{Sensors (From Top to Bottom):} Ouster OS1 64 Channel LIDAR, Velodyne VLP-32 LIDAR, Basler Ace Camera and Karmin2 Stereo Camera.}
    \label{fig: sensorsuite}
\end{figure}

In this work, we experimentally evaluate three different 3D-LIDAR camera extrinsic calibration algorithms, specifically those presented in \cite{Huang2009}, \cite{subodhIV2020} and \cite{Owens_msg-cal:multi-sensor}. Unlike \cite{Huang2009} and \cite{subodhIV2020} which are pair wise 3D-LIDAR camera calibration algorithms, \cite{Owens_msg-cal:multi-sensor} is a multi-sensor graph based optimization algorithm that jointly calibrates an arbitrary set of such sensors. We have evaluated these algorithm on the sensor suite shown in Figure \ref{fig: sensorsuite}, and have demonstrated the varying robustness of each approach to noisy sensor data. All three methods compared here use a planar target (with known physical characteristics) as the calibration object.


\section{3D-LIDAR Camera Calibration}
\subsection{The Problem}
For the pinhole camera model, the relationship between a homogeneous 3D point, ${P}^L_i$, and its homogeneous image projection $p^C_i$, is given by
\begin{equation}
    p^C_i = K[^CR_L, ^Ct_L]P^L_i.
\end{equation}

\begin{figure}
    \centering
    \includegraphics[scale=0.40]{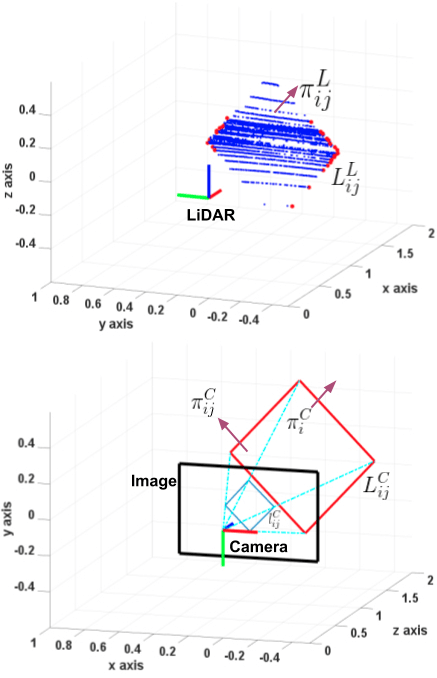}
    \caption{\textbf{Notations:} In LIDAR, the $i^{th}$ pose of the planar target yields planar points $\{P^{L}_{im}\}$ (blue, where $m=\{1, 2, ..., p_i\}$ \& $p_i$ is the number of points detected on the planar surface) and boundary points $\{Q^L_{ijn}\}$ (red, where $j=\{1, 2, 3, 4\}$ \& $n=\{1,...,q_{ij}\}$, $q_{ij}$ is the number of points on $j^{th}$ line). In Camera, the $i^{th}$ pose of the planar target yields lines ${l^{C}_{ij}}$ (where $j=\{1, 2, 3, 4\}$) and planes $\pi^{C}_{i}$ \& $\pi^{C}_{ij}$ (where $j=\{1, 2, 3, 4\}$). $\pi^{C}_{i}$ is the parameterization of the plane defined by the planar target's surface and $\pi^{C}_{ij}$ is the back-projected plane defined by the camera center and the line (edge) $l^{C}_{ij}$.}
    \label{fig: schematic}
\end{figure}

The extrinsic parameters that transform the laser coordinate system to that of the camera are captured by $[{}^CR_L, {}^Ct_L]$, where ${}^CR_L$ is the orthonormal rotation matrix and $t^C_L:=[x,y,z]^\top$ is the translation vector between the two coordinate frames.  The camera intrinsics are captured by the matrix $K$ and is assumed to be known or estimated using established monocular calibration methods (e.g., \cite{zhang-camera-calib}). The methods evaluated here require a planar target that is visible in both camera and lidar frame in order to establish the geometric constraints that allow us to estimate the rigid body transformation $[{}^CR_L, {}^Ct_L]$ between the two sensors (Figure \ref{fig: schematic}). 

\subsection{Notations}
\label{sec: notations}
We parameterize a plane in 3D as $\pi^{F} = [n^{F}_{3\times1}; d^{F}_{3\times1}]$, here $n^{F}$ is the normal to the plane in the $F$ frame of reference and $d^F$ is the vector joining the origin of the $F$ frame to the origin of the plane's frame in the $F$ frame of reference. The $i^{th}$ observation of the planar target in camera frame is parameterized as $\pi^{C}_{i} = [n^{C}_{i}; d^{C}_{i}]$, where $n^{C}_{i}$ and $d^{C}_{i}$ are the normal to the target's plane and the vector connecting the origin of the camera frame to the origin of the target frame, in the camera frame of reference. The four boundary edges (lines) of the planar target in the image are denoted as $l^{C}_{ij}$. The point of intersection $p^{C}_{ij}$ of these edges in the image plane can be easily deduced. The back-projected plane $\pi^{C}_{ij}$ associated with each line $l^{C}_{ij}$ is given by $\pi^{C}_{ij} = [K^{T}l^{C}_{ij}; \textbf{0}_{3\times1}]$ \cite{Hartley2003MVG861369} and hence, $n^{C}_{ij} = K^{T}l^{C}_{ij}$. For the $i^{th}$ pose of the planar target detected in LIDAR, we have planar points $\{P^{L}_{im}\}$ (where $m=\{1, 2, ..., p_i\}$ \& $p_i$ is the number of points detected on the planar surface) and edge points $\{Q^{L}_{ijn}\}$ (where $j=\{1, 2, 3, 4\}$ \& $n=\{1,...,q_{ij}\}$, $q_{ij}$ is the number of points on $j^{th}$ line), in the LIDAR frame of reference. $\{P^{L}_{im}\}$ can be used to estimate $\pi^{L}_i$ .

\section{3D-LIDAR Camera Extrinsic Calibration Algorithms}
In this section we describe three different 3D-LIDAR camera calibration algorithms \emph{viz.} \textbf{PPC-Cal} \cite{Huang2009} (but also implemented in \cite{unnikrishnan2005} \& \cite{PANDEY2010336}), \textbf{PBPC-Cal} \cite{subodhIV2020} and \textbf{MSG-Cal} \cite{Owens_msg-cal:multi-sensor}. We will analyze  the results of experimentally evaluating these methods in Section \ref{sec: experiments}.

\subsection{\textbf{PPC-Cal:} Point to Plane Constraint Calibration}
\label{subsec: methodA}
\textbf{PPC-Cal} has been implemented in \cite{unnikrishnan2005}, \cite{Huang2009} and \cite{PANDEY2010336}. A checkerboard pattern is printed on the planar target to facilitate the estimation of the target's plane parameters in the camera frame.  

\subsubsection{Data Collection}
\label{sec: ppcDC}
For the $i^{th}$ observation of the planar target, the points $\{P^{L}_{im}\}$ on its surface in LIDAR frame can be detected by a RANSAC\cite{RANSAC} based plane segmentation algorithm available with the Point Cloud Library (PCL) \cite{PCL} and the plane parameters $\pi^{C}_{i}$ in the Camera frame can be estimated by OpenCV's \cite{opencvlibrary} checkerboard detection module. 

Given the $i^{th}$ pose of the planar target, each $P^{L}_{im}$ and $\pi^{C}_{i}$ ($\pi^{C}_{i} = [n^{C}_{i}; d^{C}_{i}]$) pair satisfy a \textit{point to plane} constraint ( Equation \ref{eqn: point2plane}) which involves $[{}^CR_{L}, {}^Ct_{L}]$.  
\begin{equation}
    n^{C}_{i}.(^{C}R_{L}P^{L}_{im} + ^{C}t_{L} - d^{C}_{i}) = 0
    \label{eqn: point2plane}
\end{equation}
\subsubsection{Optimization}
The cost function formed by the \textit{point to plane} constraint (Equation \ref{eqn: point2plane}) is given in Equation \ref{eqn: CostFn1}.
\begin{multline}
\label{eqn: CostFn1}
P_1 = \sum_{i=1}^{M}\frac{1}{p_i}\sum_{m=1}^{p_{i}} \norm{(n^{C}_{i})^{T}(^{C}R_{L}P^{L}_{im} + ^{C}t_{L} - d^{C}_{i})}^{2}
\end{multline}

Here, $p_i$ is the number of LIDAR points lying on the planar target in the $i^{th}$ observation and $M$ is the total number of observations.
To obtain an estimate $[{}^C\Tilde{R}_L, {}^C\Tilde{t}_L]$, Equation \ref{eqn: CostFn1} needs to be minimized with respect to $[{}^CR_L, {}^Ct_L]$ which involves solving a minimization problem given in Equation \ref{eqn: CostFn1Minimization}, formed by Equation \ref{eqn: CostFn1} .

\begin{equation}
\label{eqn: CostFn1Minimization}
    [{}^C\Tilde{R}_L, {}^C\Tilde{t}_L] = \argmin_{[{}^CR_L, {}^Ct_L]} P_1
\end{equation}


We use a non-linear least square optimization library, Ceres \cite{ceres-solver}, to solve the minimization problem given in Equation \ref{eqn: CostFn1Minimization}. As mentioned in \cite{PANDEY2010336}, we need at-least 3 non-co-planar views to solve the optimization problem formed by Equation \ref{eqn: CostFn1Minimization} but in practice it is advisable to collect numerous observations to better constrain the optimization. We used about 30 observations in the experiments for each LIDAR camera pair.

\subsubsection{Remarks}
\textbf{PPC-Cal} requires only the planar points in LIDAR frame and plane parameters in camera frame to estimate $[^CR_L, ^Ct_L]$. Therefore, data collection is easy and fast. Additionally, the use of the checkerboard further accelerates the process of plane detection in camera frame.

\subsection{\textbf{PBPC-Cal:} Point to Back-projected Plane Constraint Calibration}
\label{subsec: methodB}
\textbf{PBPC-Cal} has been implemented in \cite{subodhIV2020} for 3D-LIDAR camera calibration. In addition to the \textit{point to plane} constraint (Equation \ref{eqn: point2plane}) used in \textbf{PPC-Cal}, \textbf{PBPC-Cal} uses a \textit{point to back projected plane} constraint (Equation \ref{eqn: point2backprojplane}). This method requires detection of not only the plane but also the edges of the planar target, in both sensing modalities. 

\subsubsection{Data Collection}
The planar $\{P^{L}_{im}\}$ and edge points $\{Q^{L}_{ijn}\}$ in LIDAR frame are detected using RANSAC based respective plane and line segmentation algorithms available in PCL. The plane parameters $\pi^{C}_{i}$ and the edge parameters $l^{C}_{ij}$ are estimated in the Camera frame using OpenCV. The edge detection in the camera frame is done using the Line Segment Detector (LSD) \cite{LSD4731268} available in OpenCV. Unlike \textbf{PPC-Cal}, this approach does not use a checkerboard for detection of the planar target in camera frame. It rather uses the points of intersection $p^{C}_{ij}$ of the detected edges $l^{C}_{ij}$ of the planar target in the image and the known physical dimensions of the calibration target to solve a Perspective-n-Point (PnP) algorithm and estimate the plane parameters $\pi^{C}_{i}$ in the camera frame. 

Given the $i^{th}$ pose of the planar target, each $Q^{L}_{ijn}$ and $l^{C}_{ij}$ pair satisfy the \textit{point to back projected plane} constraint (Equation \ref{eqn: point2backprojplane}) which involves $[{}^CR_{L}, {}^Ct_{L}]$.
\begin{equation}
    n^{C}_{ij}.(^{C}R_{L}Q^{L}_{ijn} + ^{C}t_{L}) = 0
    \label{eqn: point2backprojplane}
\end{equation}
Where $n^{C}_{ij} = K^{T}l^{C}_{ij}$ is the normal to the back-projected plane formed by the camera center and the line $l^{C}_{ij}$ (Fig \ref{fig: schematic}) and $K$ is the camera instrinsic matrix.

\subsubsection{Optimization}
The cost function formed by \textit{point to back projected plane} constraint (Equation \ref{eqn: point2backprojplane}) is given in Equation \ref{eqn: CostFn2}.
\begin{multline}
\label{eqn: CostFn2}
P_2 = \sum_{i=1}^{N} \sum_{j=1}^{4} \frac{1}{q_{ij}} \sum_{n=1}^{q_{ij}} \norm{(n^{C}_{ij})^{T}(^{C}R_{L}Q^{L}_{ijn} + ^{C}t_{L})}^{2}
\end{multline}

Here $q_{ij}$ is the number of points lying on the $j^{th}$ line in the $i^{th}$ observation and $N$ is the number of observations. To obtain an estimate $[{}^C\Tilde{R}_L, {}^C\Tilde{t}_L]$, Equation \ref{eqn: CostFn2} needs to be minimized with respect to $[{}^CR_L, {}^Ct_L]$ which involves solving a minimization problem given in Equation \ref{eqn: CostFn2Minimization}, formed by Equation \ref{eqn: CostFn2} .
\begin{equation}
\label{eqn: CostFn2Minimization}
    [{}^C\Tilde{R}_L, {}^C\Tilde{t}_L] = \argmin_{[{}^CR_L, {}^Ct_L]} P_2
\end{equation}
In this method, the minimization problem given in Equation \ref{eqn: CostFn1Minimization} is solved first and its solution is used to initialize the minimization problem given in Equation \ref{eqn: CostFn2Minimization}. Like \textbf{PPC-Cal}, the Ceres Solver \cite{ceres-solver} is used. The \textit{point to back-projected plane} constraint given in Equation \ref{eqn: point2backprojplane} is equivalent to the line correspondence equation given in \cite{Hartley2003MVG861369} (2004, p. 180). The solution to such a system of equation is given by the DLT-Lines method and requires at least 6 noise free line correspondences \cite{PribylZC16a} between the LIDAR and camera views. Since the planar target has 4 sides , theoretically, we need at least 2 distinct views to solve this system but use of several frames is advised. We used about 30 observations in the experiments for each LIDAR camera pair.

\subsubsection{Remarks}
As compared to \textbf{PPC-Cal}, \textbf{PBPC-Cal} requires both planar points and the points lying on the edges of the target, in the LIDAR frame. Data collection is tedious because successful detection of all edges in LIDAR pointcloud depends on the way the target is held. This method requires the target to be held in a diagonal sense as shown in Figures \ref{fig: comparisonABC} and \ref{fig: vlp_left_withAndwithout_ouster} such that any edge of the target is not parallel to the scan lines of the LIDAR. Since this method doesn't use any fiducial marker like checkerboard or ArUco or AprilTag, the detection of planar target in image depends on OpenCV's Line Segment Detector (LSD) which may be affected by illumination. Moreover some heuristics are necessary to establish association between lines in the image and corresponding points in the pointcloud.
\subsection{\textbf{MSG-Cal:} Multi-Sensor Graph based Calibration}
\textbf{PPC-Cal} and \textbf{PBPC-Cal} do pair-wise calibration of a 3D-LIDAR and camera system but \textbf{MSG-Cal} described in \cite{Owens_msg-cal:multi-sensor} adds another layer over pair-wise calibration of sensors by utilizing a graph based optimization paradigm to jointly calibrate several sensors. The first step involves pair-wise calibration of all the sensors present in the sensor suite and the second step involves a global optimization using g2o \cite{g2o}, a general framework for graph optimization.

\textbf{PPC-Cal} and \textbf{PBPC-Cal} described previously can only cross calibrate 3D-LIDARs and cameras but \textbf{MSG-Cal} can cross calibrate across all pair-wise sensing modalities\footnote{\emph{i.e.} 3D-LIDAR$\leftrightarrow$3D-LIDAR, 3D-LIDAR$\leftrightarrow$Camera, 3D-LIDAR$\leftrightarrow$2D-LIDAR, 2D-LIDAR$\leftrightarrow$Camera \& Camera$\leftrightarrow$Camera} except for a 2D-LIDAR with 2D-LIDAR, and can jointly calibrate any configuration of 3D-lidars, cameras, and 2D-lidars.

\subsubsection{Data Collection}
For LIDAR pointcloud, \textbf{MSG-Cal} uses PCL to make a model of the environment using the first frame (with no calibration target present), and when the target is introduced into the environment in subsequent frames, it is detected by background subtraction from the pre-built model. The result of background subtraction gives a dominant plane and many other points which may be sparse and random. With simple heuristics such as density of points and approximate size of the target, it is easy to filter out the dominant plane ($\pi^{L} = [n^{L}; d^{L}]$). An AprilTag pattern is used for detection of the planar target ($\pi^{C} = [n^{C}; d^{C}]$) in camera. 

\subsubsection{Pair-wise Calibration}
\label{sec: msgcalpair-wise}
For 3D-LIDAR$\leftrightarrow$camera, 3D-LIDAR$\leftrightarrow$3D-LIDAR and camera$\leftrightarrow$camera calibration the constraints are given by Equation \ref{eqn: normal2normal} and Equation \ref{eqn: modifiedpoint2plane}. For the $i^{th}$ observation, the normal alignment constraint is given by Equation \ref{eqn: normal2normal}
\begin{equation}
\label{eqn: normal2normal}
    n^{C}_{i} - {}^CR_{L}n^{L}_{i} = 0
\end{equation}
Then, a point lying on a planar surface satisfies Equation \ref{eqn: lidarPoint2Plane}.
\begin{equation}
\label{eqn: lidarPoint2Plane}
    n^{L}_{i}.(P^{L}_{im} - d^{L}_i) = 0
\end{equation}
Using Equation \ref{eqn: normal2normal} and Equation \ref{eqn: lidarPoint2Plane} in Equation \ref{eqn: point2plane} we have
a modified version of the \textit{point to plane} constraint (which can be called a \textit{plane to plane} constraint as $P^{L}_{im}$ has been eliminated),
\begin{align}
\label{eqn: modifiedpoint2plane}
     n^{C}_{i} . {}^{C}t_{L} + n^{L}_{i} . d^{L}_i - n^{C}_{i} . d^{C}_i = 0
\end{align}
Estimation of pair-wise SE(3) transformation parameters for \textit{plane to plane} correspondences across sensors is done by minimizing a joint cost function (Equation \ref{eqn: JointCostFunc}) formed by Equation \ref{eqn: normal2normal} and \ref{eqn: modifiedpoint2plane},

\begin{multline}
\label{eqn: JointCostFunc}
P_3=\sum_{i=1}^{M}\norm{(n^{C}_{i} - {}^CR_{L}n^{L}_{i})}^{2}+ \\ \sum_{i=1}^{M}\norm{(n^{C}_{i} . ^{C}t_{L} + n^{L}_{i} . d^{L}_i - n^{C}_{i} . d^{C}_i)}^{2},
\end{multline}
where $M$ is the number of observations. The minimization problem is given in Equation \ref{eqn: CostFn3Minimization}.
\begin{equation}
\label{eqn: CostFn3Minimization}
    [{}^C\Tilde{R}_L, {}^C\Tilde{t}_L] = \argmin_{[{}^CR_L, {}^Ct_L]} P_3
\end{equation}

\subsubsection{Global Calibration}
In this phase, a hypergraph composed of several node and edge types that exploit the pair-wise relative transforms as an initialization for the global sensor pose graph is constructed. The goal of the global graph approach is to incorporate all the information into a unified optimization structure, requiring a single optimization run to calibrate many sensors. The sensor poses are the unknowns that are estimated simultaneously in a global frame. In contrast to \textbf{PPC-Cal} and \textbf{PBPC-Cal}, \textbf{MSG-Cal} incorporates new global graph constraints for camera$\leftrightarrow$camera sensor pairs that incorporate the positions of individual AprilTags seen by multiple cameras. We collected about 100 observations for the experiments to ensure all the sensor pairs have sufficient detections. 
\subsubsection{Remarks}
Like \textbf{PPC-Cal}, \textbf{MSG-Cal} needs only points lying on the planar target in LIDAR frame and uses an AprilTag for easy detection of the planar target in camera frame which makes data collection relatively easy. 

\section{System Description}
Our sensor suite (Figure \ref{fig: sensorsuite}) consists of an Ouster OS1 64 Channel LIDAR, a Velodyne VLP-32 LIDAR, a Basler Ace camera $[1600 \times 1200]$ and the Karmin2 Stereo Vision System (which comprises two Basler Cameras $[800 \times 600]$) such that the factory stereo calibration is known.

\section{Experiments and Results}
\label{sec: experiments}

\begin{figure*}[!ht]
\centering
    \includegraphics[width=\textwidth]{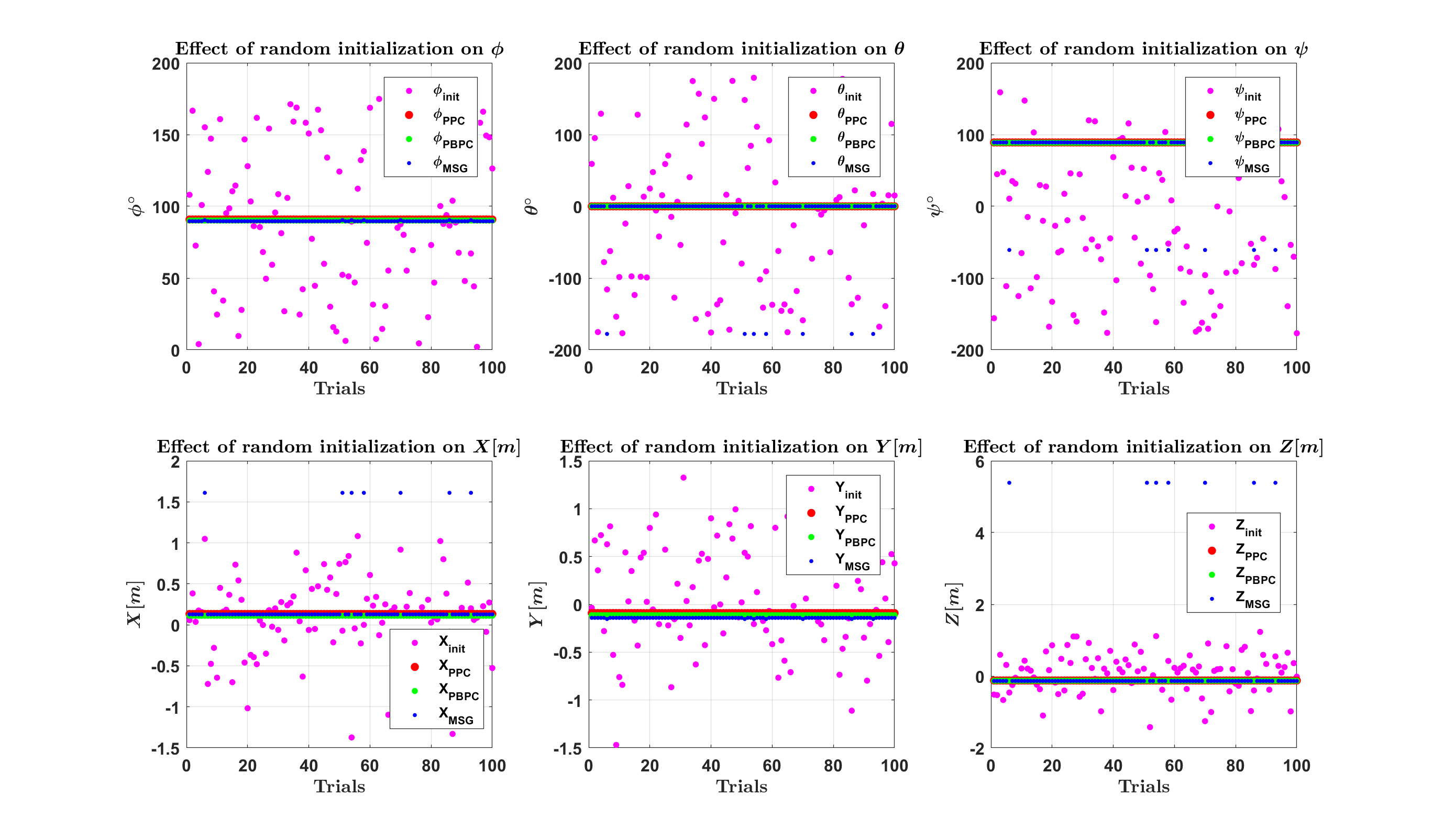}
    \caption{Comparing performance of \textbf{PPC-Cal}, \textbf{PBPC-Cal} and \textbf{MSG-Cal} to random initialization. This figure shows the calibration result for the left stereo camera and Velodyne VLP-32 LIDAR under random initialization}
    \label{fig: randomInitialization}
\end{figure*}

\begin{figure*}[!ht]
  \centering
  \subfloat[MLRE = \textcolor{orange}{2.94664} with \textbf{PPC-Cal} for VLP-32 and Right Stereo Camera]{\includegraphics[width=0.14\textwidth]{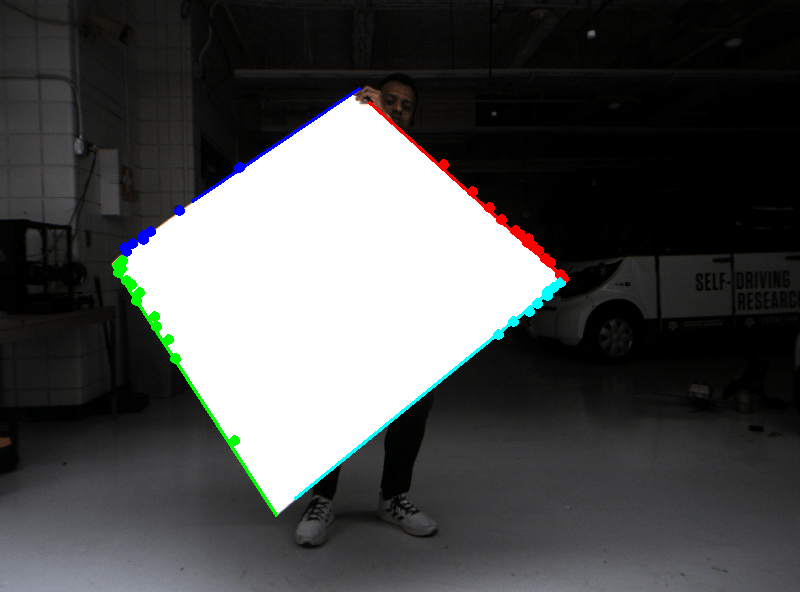}\label{fig: vlp_right_methodA}}
  \quad
  \subfloat[MLRE = \textcolor{blue}{1.88719} with \textbf{PBPC-Cal} for VLP-32 and Right Stereo Camera]{\includegraphics[width=0.14\textwidth]{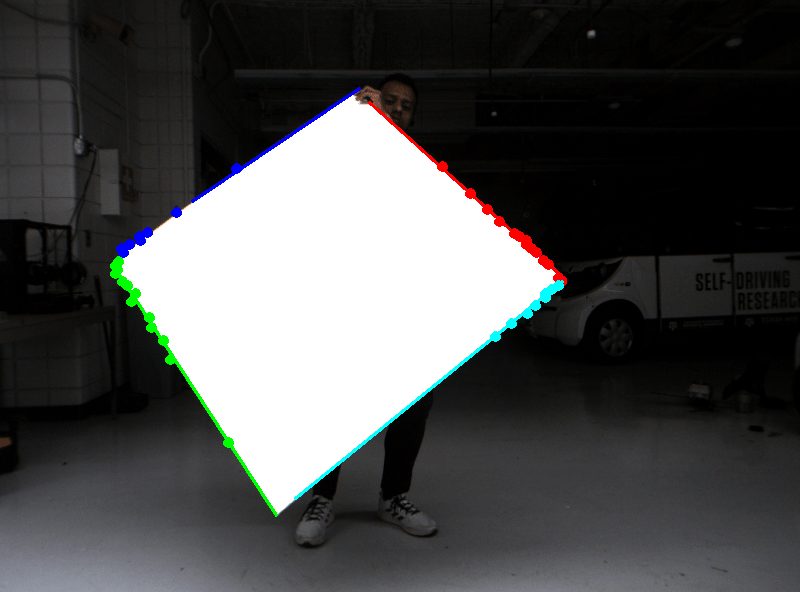}\label{fig: vlp_right_methodB}}
  \quad
    \subfloat[MLRE = \textcolor{red}{11.3415} with \textbf{MSG-Cal} for VLP-32 and Right Stereo Camera]{\includegraphics[width=0.14\textwidth]{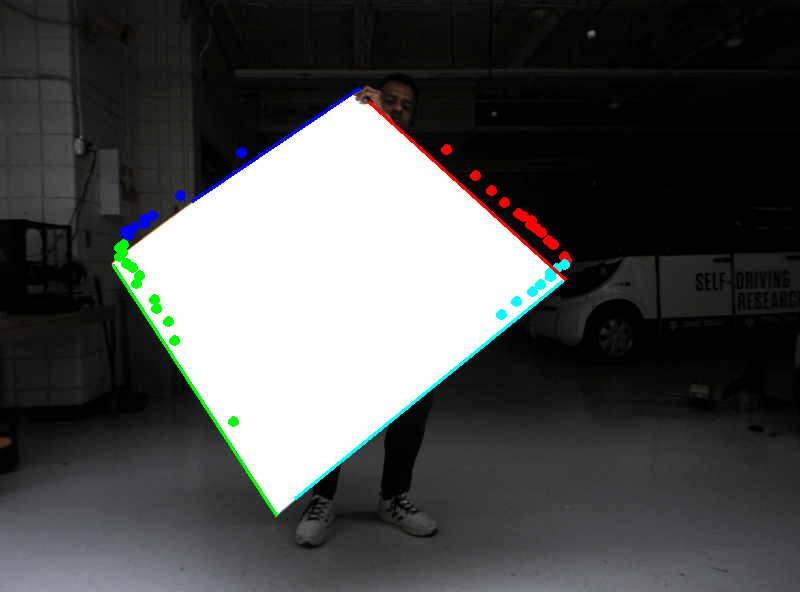}\label{fig: vlp_right_methodC}} 
  \quad
  \subfloat[MLRE = \textcolor{orange}{5.63206} with \textbf{PPC-Cal} for OS1-64 and Right Stereo Camera]{\includegraphics[width=0.14\textwidth]{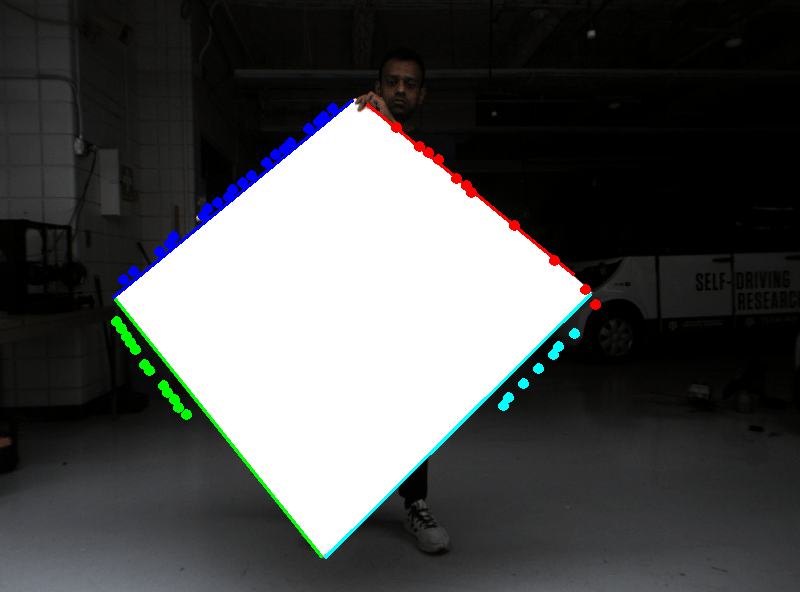}\label{fig: os_right_methodA}}
  \quad
  \subfloat[MLRE = \textcolor{blue}{1.74138} with \textbf{PBPC-Cal} for OS1-64 and Right Stereo Camera]{\includegraphics[width=0.14\textwidth]{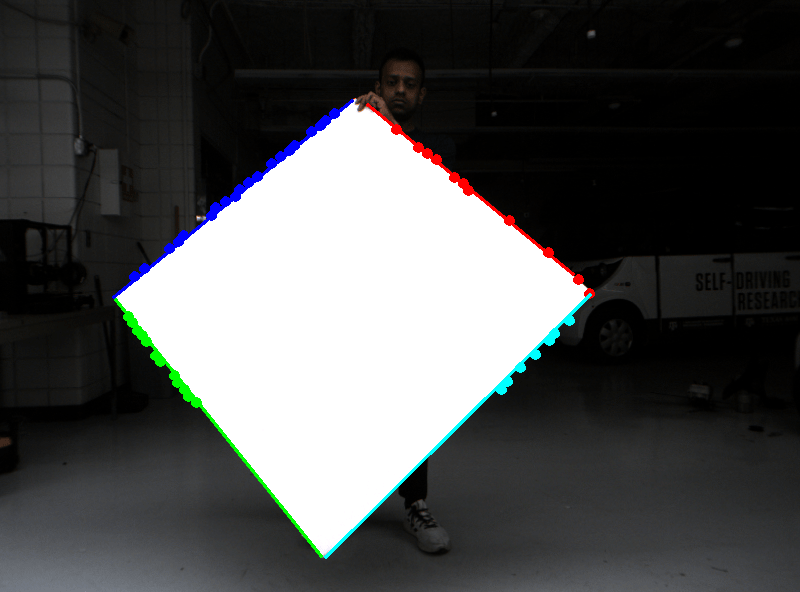}\label{fig: os_right_methodB}}
  \quad
    \subfloat[MLRE = \textcolor{red}{11.0735} with \textbf{MSG-Cal} for OS1-64 and Right Stereo Camera]{\includegraphics[width=0.14\textwidth]{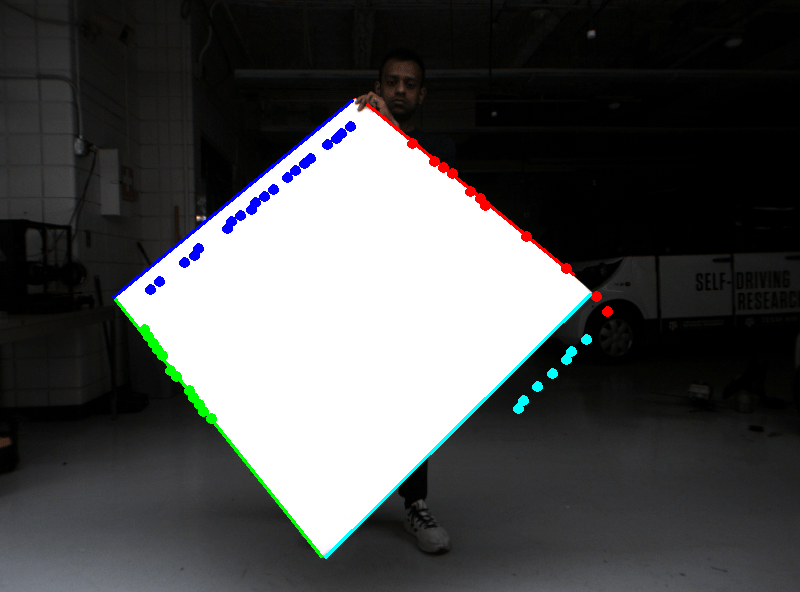}\label{fig: os_right_methodC}} \\
  \subfloat[MLRE = \textcolor{orange}{2.21383} with \textbf{PPC-Cal} for VLP-32 and Basler Ace Camera]{\includegraphics[width=0.14\textwidth]{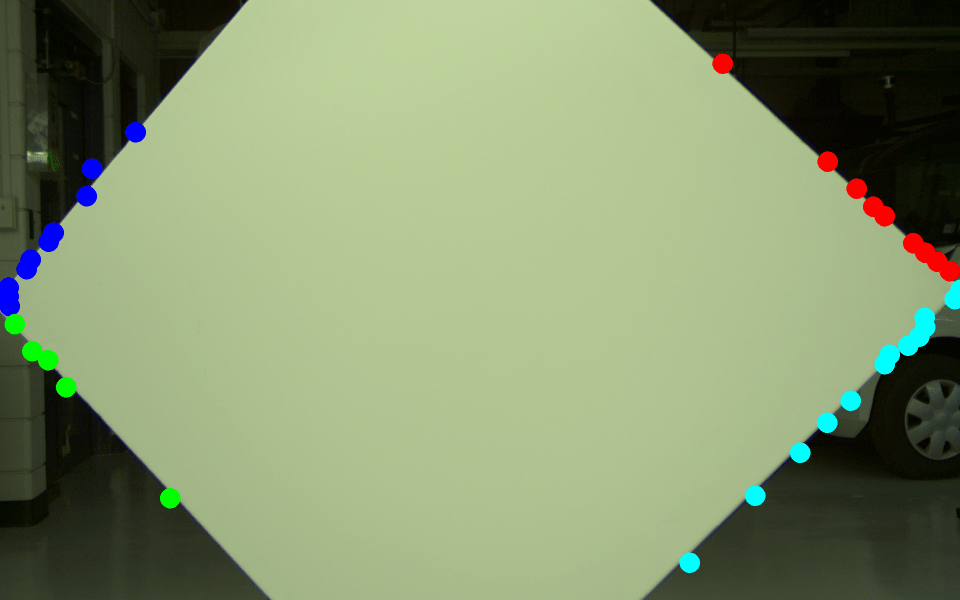}\label{fig: vlp_basler_methodA}}
  \quad
  \subfloat[MLRE = \textcolor{blue}{1.9423} with \textbf{PBPC-Cal} for VLP-32 and Basler Ace Camera]{\includegraphics[width=0.14\textwidth]{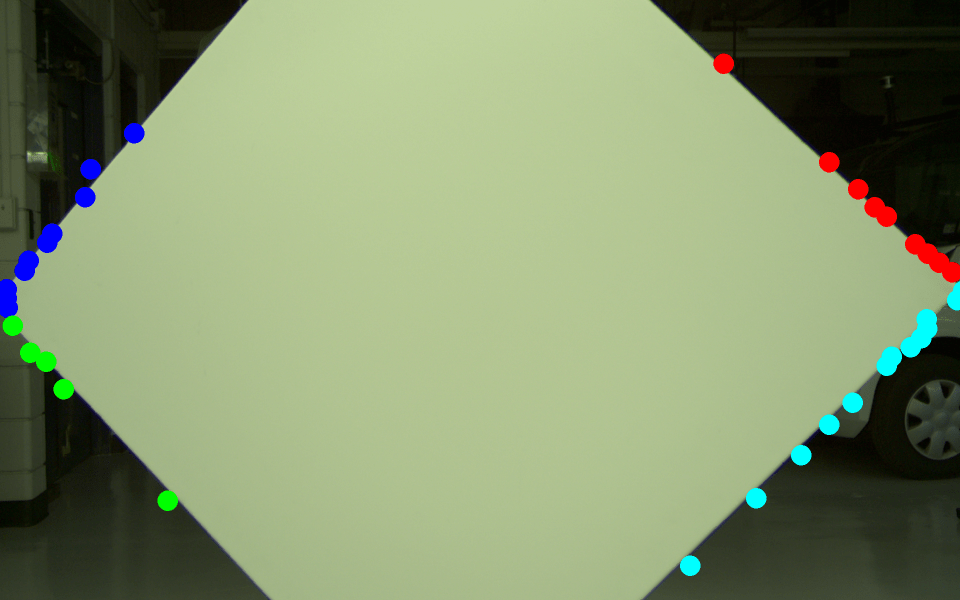}\label{fig: vlp_basler_methodB}}
  \quad
 \subfloat[MLRE = \textcolor{red}{8.8254} with \textbf{MSG-Cal} for VLP-32 and Basler Ace Camera]{\includegraphics[width=0.14\textwidth]{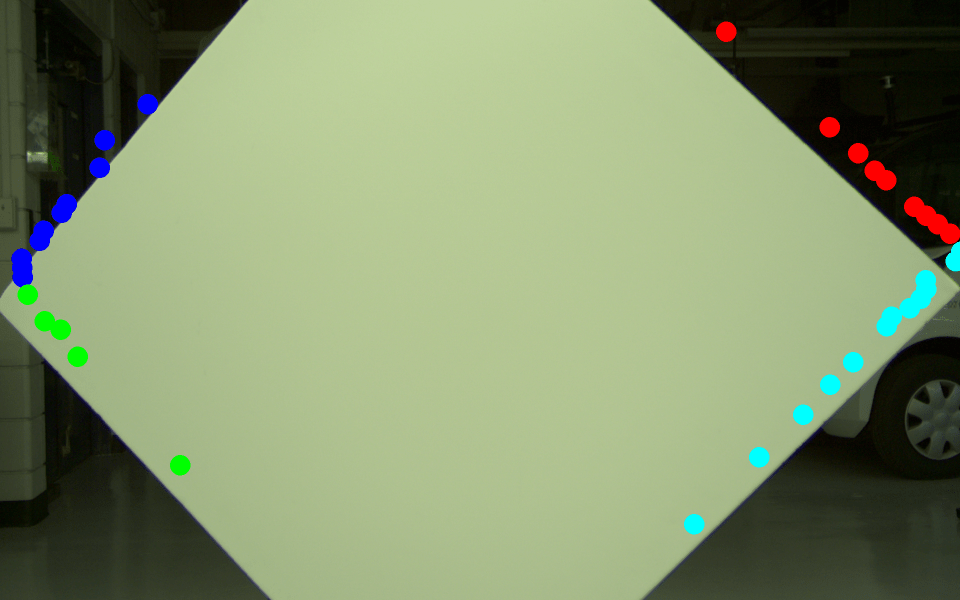}\label{fig: vlp_basler_methodC}}
    \quad
  \subfloat[MLRE = \textcolor{orange}{6.95193} with \textbf{PPC-Cal} for OS1-64 and Basler Ace Camera]{\includegraphics[width=0.14\textwidth]{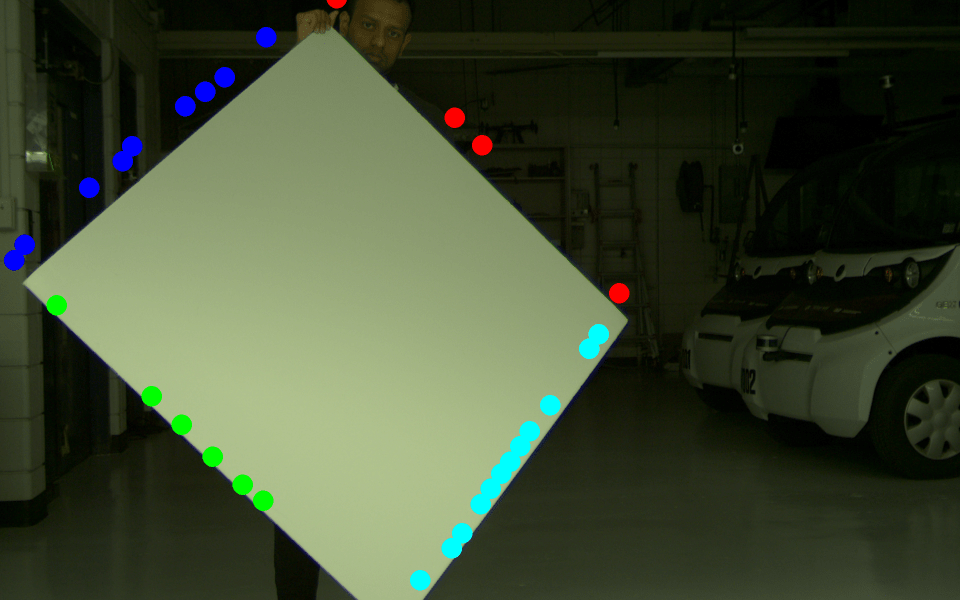}\label{fig: os_basler_methodA}}
  \quad
  \subfloat[MLRE = \textcolor{blue}{1.80414} with \textbf{PBPC-Cal} for OS1-64 and Basler Ace Camera]{\includegraphics[width=0.14\textwidth]{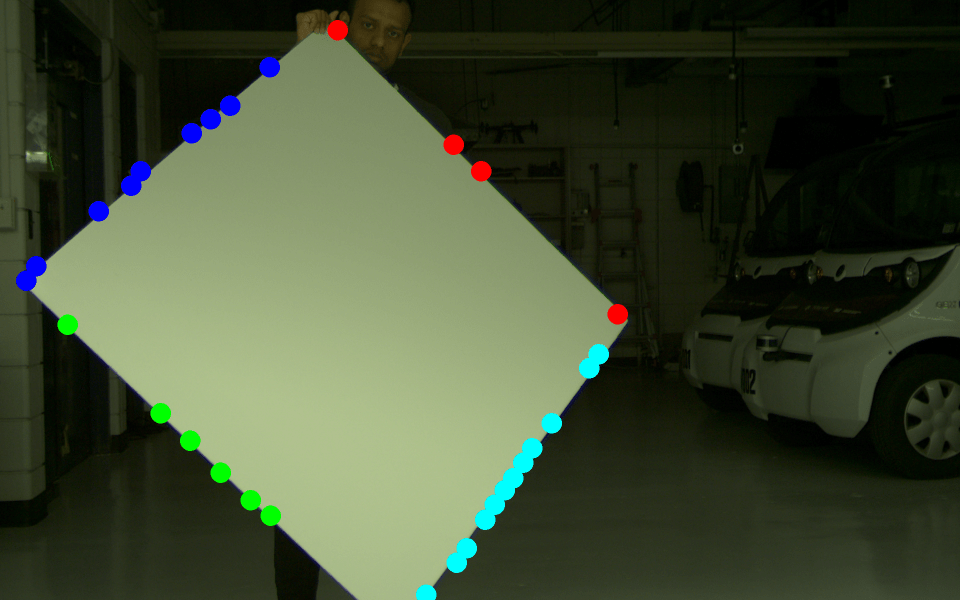}\label{fig: os_basler_methodB}}
  \quad
 \subfloat[MLRE = \textcolor{red}{11.7995} with \textbf{MSG-Cal} for OS1-64 and Basler Ace Camera]{\includegraphics[width=0.14\textwidth]{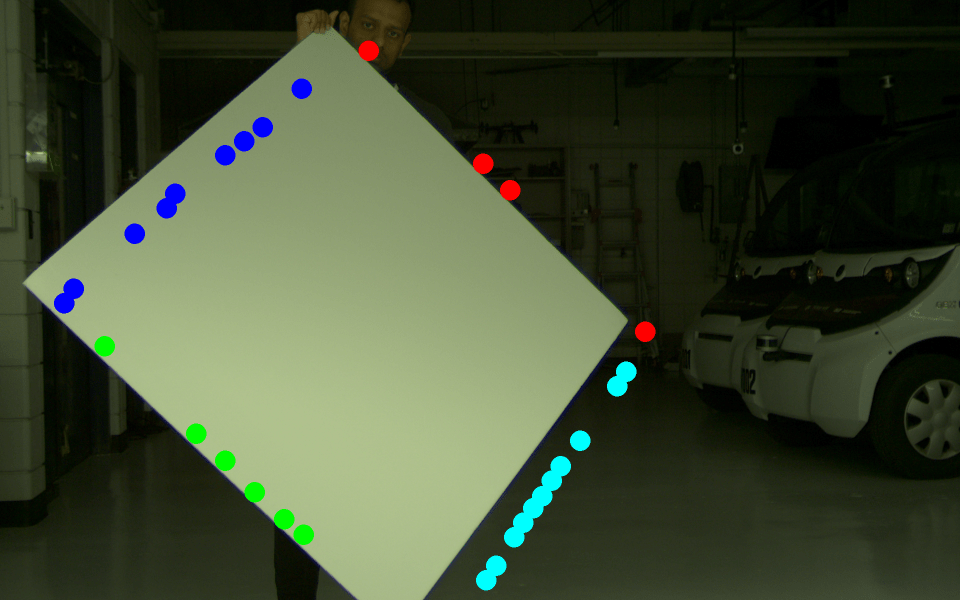}\label{fig: os_basler_methodC}} 
    
  \caption{Comparing performance of \textbf{PPC-Cal}, \textbf{PBPC-Cal} and \textbf{MSG-Cal} using Mean Line Re-projection Error}
  \label{fig: comparisonABC}
\end{figure*}

We compare the performance \footnote{\textcolor{blue}{blue: best performance}, \textcolor{red}{red: worst performance}} of \textbf{PPC-Cal}, \textbf{PBPC-Cal} and \textbf{MSG-Cal} by using these methods to calibrate our sensor suite (Figure \ref{fig: sensorsuite}). First, we evaluate their robustness to random initial conditions (Figure \ref{fig: randomInitialization}) drawn from a zero mean normal distribution with standard deviation of $90^\circ$ and $50$ cm for rotation and translation respectively. We notice that \textbf{PPC-Cal} and \textbf{PBPC-Cal} are robust to initialization while \textbf{MSG-Cal} exhibits divergence in a few cases, in which the optimization arrives at the same incorrect local minima. Besides, we notice that all the methods converge to nearly the same rotational values but show variation in translation values. This is because the \textit{point to plane} constraint (which is used in all the three methods) is good at constraining rotation but it needs several observations to constrain translation. The \textit{point to back-projected plane} constraint (used in \textbf{PBPC-Cal}) helps translation estimation accuracy by providing additional constraints at each measurement. While we don't expect such bad initial guesses in practice (and initial guesses of Identity converged for each algorithm across multiple datasets), we are effectively showing how well each formulation constrains the optimization. Since the minimization problem(s) (Equation \ref{eqn: CostFn1Minimization}, \ref{eqn: CostFn2Minimization}, \ref{eqn: CostFn3Minimization}) solved to estimate $[^CR_L, ^Ct_L]$ are highly non-linear and involve parameters on manifolds, the convergence over several random initialization assures the user that the calibration process can be executed with any initial guess.

In the absence of ground truth we verify our algorithms by using the estimated parameters \begin{enumerate*} [label=\itshape\alph*\upshape)]
\item to compare it against the factory stereo calibration \footnote{We use the estimated $T^{C_1}_L$ and $T^{C_2}_L$ and compare $T^{C_1}_{L}(T^{C_2}_{L})^{-1}$ with the given factory stereo calibration $T^{C_1}_{C_2}$} and
\item to project points lying on the edges of the planar target in LIDAR frame on the Camera image and calculate the mean line re-projection errors (MLRE).\footnote{MLRE is the average $\bot$ distance between $\{l^{C}_{ij}\}$ and $\{Q^{L}_{ijn}\}$ projected on the image using the estimated $[{}^CR_{L}, {}^Ct_{L}]$}
\end {enumerate*} MLRE is an independent evaluation metric since none of the methods we compare in this work use it as a residual in their respective optimizations. 

\begin{table}[!ht]
\setlength{\tabcolsep}{2pt}
\centering
\begin{tabular}{| c | c | c | c |} 
\hline
& \textbf{PPC-Cal} & \textbf{PBPC-Cal} & \textbf{MSG-Cal} \\ \hline
$\alpha^{\circ}_{err}$ &  \textcolor{blue}{-0.0055535}  &  \textcolor{red}{0.19277}  & 0.10103 \\ \hline
$\beta^{\circ}_{err}$ &  0.097271  &  \textcolor{red}{0.19995}  & \textcolor{blue}{0.057334} \\ \hline
$\gamma^{\circ}_{err}$ &  \textcolor{blue}{-0.081701}  &  \textcolor{red}{-0.12867}  & -0.11242 \\ \hline
$X_{err}$ $[m]$ &  0.00304  &  \textcolor{red}{0.00640}  & \textcolor{blue}{-0.00113} \\ \hline
$Y_{err}$ $[m]$ &  \textcolor{red}{-0.00439}  &  \textcolor{blue}{-0.00352}  & -0.00377 \\ \hline
$Z_{err}$ $[m]$ &  \textcolor{red}{0.01124}  &  \textcolor{blue}{0.00459}  & 0.01072 \\ \hline
\end{tabular}
\caption{Errors with respect to factory stereo calibration for Velodyne VLP-32 LIDAR and the stereo rig.}
\label{table: stereo_VLP}
\end{table}


\begin{table}[!ht]
\setlength{\tabcolsep}{2pt}
\centering
\begin{tabular}{| c | c | c | c |} 
\hline
\multicolumn{4}{|c|}{Errors with respect to factory stereo calibration}\\ \hline
& \textbf{PPC-Cal} & \textbf{PBPC-Cal} & \textbf{MSG-Cal} \\ \hline
$\alpha^{\circ}_{err}$ &  \textcolor{red}{0.51756}  &  \textcolor{blue}{0.068189}  & 0.10103 \\ \hline
$\beta^{\circ}_{err}$ &  \textcolor{blue}{0.037753}  &  \textcolor{red}{0.12717}  & 0.057334 \\ \hline
$\gamma^{\circ}_{err}$ &  \textcolor{blue}{-0.061076}  &  \textcolor{red}{-0.22650}  & -0.11242 \\ \hline
$X_{err}$ $[m]$ &  \textcolor{blue}{-0.00101}  &  \textcolor{red}{-0.00507}  & -0.00113 \\ \hline
$Y_{err}$ $[m]$ &  \textcolor{blue}{-0.00102}  &  \textcolor{red}{-0.00622}  & -0.00377 \\ \hline
$Z_{err}$ $[m]$ &  0.00877  &  \textcolor{blue}{0.00475}  & \textcolor{red}{0.01072} \\ \hline
\end{tabular}
\caption{Errors with respect to factory stereo calibration for Ouster 64 Channel LIDAR and the stereo rig}
\label{table: stereo_os}
\end{table}

\begin{table}[!ht]
\setlength{\tabcolsep}{2.25pt}
\centering
\begin{tabular}{| c | c | c | c |} 
\hline
  \textbf{3D-LIDAR Camera Pair} & \multicolumn{3}{|c|}{MLRE}\\ 
 \hline
  & \textbf{PPC-Cal} & \textbf{PBPC-Cal} & \textbf{MSG-Cal} \\ [0.5ex] 
 \hline
 \texttt{VLP-32 $\leftrightarrow$ Stereo Left} & 2.57316 & \textcolor{blue}{1.94707} & \textcolor{red}{11.6547} \\ \hline
 \texttt{VLP-32 $\leftrightarrow$ Stereo Right} & 2.94664 & \textcolor{blue}{1.88719} & \textcolor{red}{11.3415} \\ \hline
 \texttt{OS1 $\leftrightarrow$ Stereo Left} & 5.51552 & \textcolor{blue}{1.76985} & \textcolor{red}{10.3954} \\ \hline
 \texttt{OS1 $\leftrightarrow$ Stereo Right} & 5.63206 & \textcolor{blue}{1.74138} & \textcolor{red}{11.0735} \\ \hline
 \texttt{VLP-32 $\leftrightarrow$ Basler} & 2.21383 & \textcolor{blue}{1.9423} & \textcolor{red}{8.8254} \\ \hline
 \texttt{OS1 $\leftrightarrow$ Basler} & 6.95193 & \textcolor{blue}{1.80414} & \textcolor{red}{11.7995} \\ \hline
 \texttt{Standard Deviation} & \textcolor{red}{1.9723} & \textcolor{blue}{0.089039} & \textcolor{black}{1.1087} \\ \hline
\end{tabular}
\caption{MLRE (in pixels) for various 3D-LIDAR Camera Pairs with \textbf{PPC-Cal}, \textbf{PBPC-Cal} and \textbf{MSG-Cal}}
\label{table: line_rep_err_1}
\end{table}

In Table \ref{table: stereo_VLP} we can see that \textbf{PPC-Cal} and \textbf{MSG-Cal} show error in the order of 1 cm along the stereo baseline dimension (Z axis) as compared to 4.5 mm in \textbf{PBPC-Cal}. In Table \ref{table: stereo_os}, \textbf{PPC-Cal} shows better performance than the others. We can see that \textbf{MSG-Cal} gives the same error in both the Tables \ref{table: stereo_VLP} \& \ref{table: stereo_os}. It is so because \textbf{MSG-Cal} is a graph based approach which does joint optimization of all the sensors together and also does camera$\leftrightarrow$camera pair-wise calibration. It is difficult to draw definitive conclusions by comparing only the stereo errors. Hence, we proceed to compare the MLRE in Table \ref{table: line_rep_err_1} and Figure \ref{fig: comparisonABC}.

\begin{figure}[!ht]
  \centering
  \subfloat[MLRE = \textcolor{red}{11.6547} with Ouster Lidar in the sensor suite for VLP-32 and Left Stereo Camera]{\includegraphics[width=0.14\textwidth]{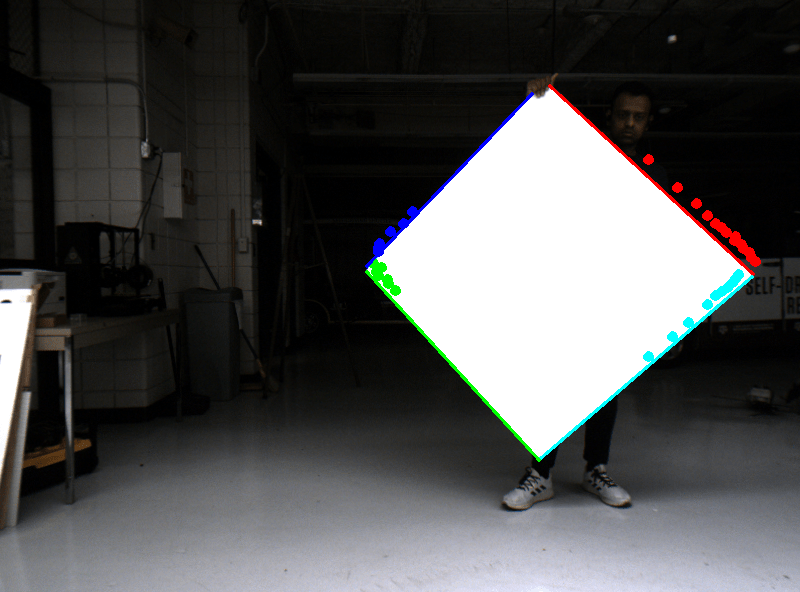}\label{fig: vlp_left_with_ouster}}
  \quad
  \subfloat[MLRE = \textcolor{red}{11.3415} with Ouster Lidar in the sensor suite for VLP-32 and Right Stereo Camera]{\includegraphics[width=0.14\textwidth]{figures/vlp_right_methodC.png}\label{fig: vlp_right_methodC}}
  \quad
    \subfloat[MLRE = \textcolor{red}{8.8254} with Ouster Lidar in the sensor suite for VLP-32 and Basler Ace Camera]{\includegraphics[width=0.14\textwidth]{figures/vlp_basler_methodC.png}\label{fig: vlp_basler_methodC}}\\
  \subfloat[MLRE = \textcolor{blue}{3.25161} without Ouster Lidar in the sensor suite for VLP-32 and Left Stereo Camera]{\includegraphics[width=0.14\textwidth]{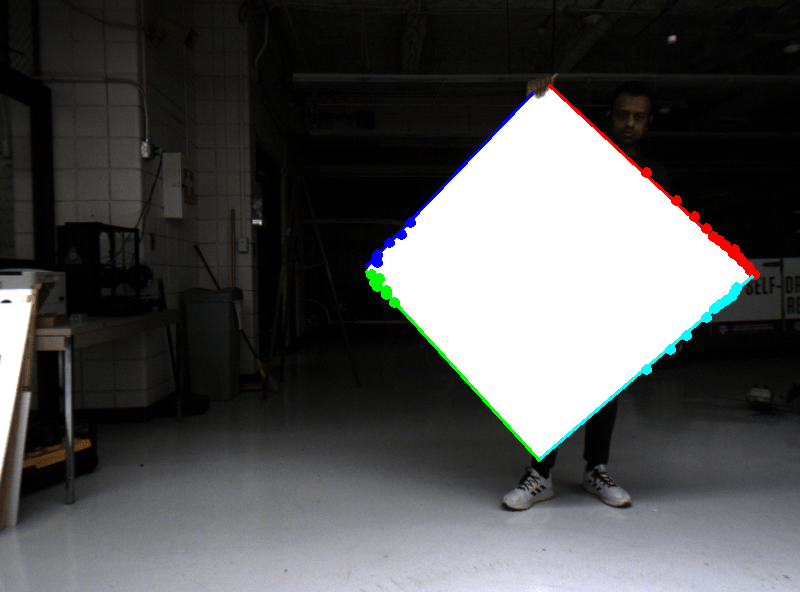}\label{fig: vlp_left_no_ouster}}
  \quad
  \subfloat[MLRE = \textcolor{blue}{2.77772} without Ouster Lidar in the sensor suite for VLP-32 and Right Stereo Camera]{\includegraphics[width=0.14\textwidth]{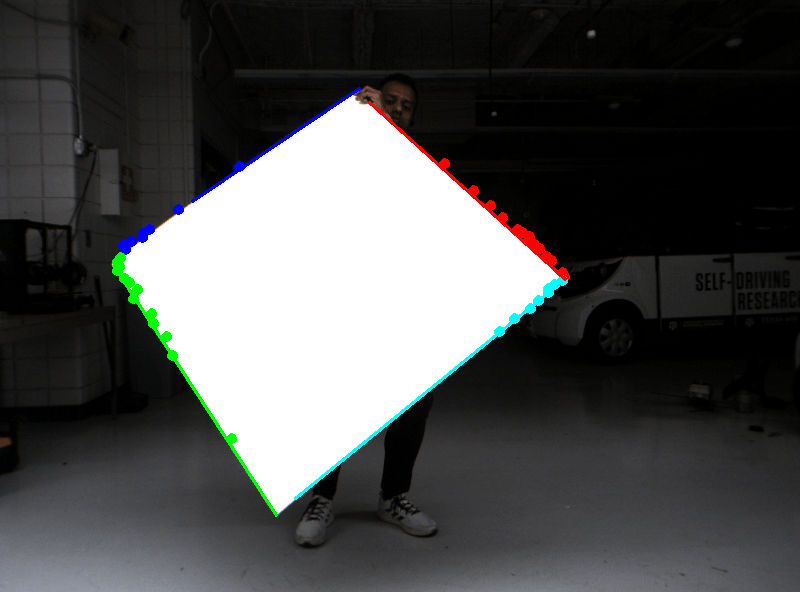}\label{fig: vlp_right_methodC_no_ouster}}
  \quad
  \subfloat[MLRE = \textcolor{blue}{2.5893} without Ouster Lidar in the sensor suite for VLP-32 and Basler Ace Camera]{\includegraphics[width=0.14\textwidth]{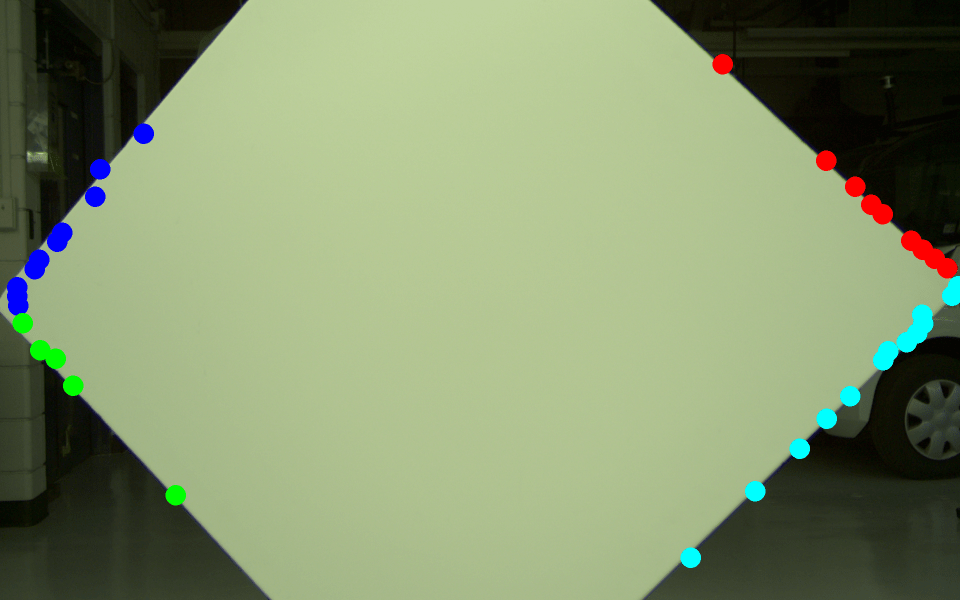}\label{fig: vlp_basler_methodC_no_ouster}}
  \caption{Comparing performance of \textbf{MSG-Cal} with (Figure \ref{fig: vlp_left_with_ouster}, Figure \ref{fig: vlp_right_methodC}, Figure \ref{fig: vlp_basler_methodC}) and without (Figure \ref{fig: vlp_left_no_ouster}, Figure \ref{fig: vlp_right_methodC_no_ouster}, Figure \ref{fig: vlp_basler_methodC_no_ouster}) Ouster LIDAR in the sensor suite(Figure \ref{fig: sensorsuite})}
  \label{fig: vlp_left_withAndwithout_ouster}
\end{figure}

From Table \ref{table: line_rep_err_1} it can be concluded that the \textbf{PBPC-Cal} performs best among all the three methods. If we compare \textbf{PPC-Cal} and \textbf{PBPC-Cal} we can see that the result of \textbf{PBPC-Cal} is consistent for all the sensor pairs, as expressed by a low standard deviation (\textcolor{blue}{0.089039} pixels) but \textbf{PPC-Cal} greater variation as evident from a high standard deviation (\textcolor{red}{1.9723} pixels). As discussed in \cite{subodhIV2020}, the Ouster LIDAR is a noisy sensor and \textbf{PPC-Cal} doesn't perform well when the Ouster Lidar is used. \textbf{MSG-Cal} produced consistent results when using various M-estimators such as Huber and Tukey cost functions, as well as testing various confidence parameters of the Ouster sensor. The confidence value of a sensor propagates to both the pair-wise and global calibration steps. For a pair-wise calibration, the RANSAC inlier threshold is scaled according to the inverse sum of each sensor's confidence, so that a more permissive threshold is provided for noisier sensors. For the graph calibration, the uncertainty associated with observations from noisier sensors is increased exponentially compared to more accurate sensors. We can hypothesize that the graph based approach \textbf{MSG-Cal} which does joint optimization will have all its nodes affected by Ouster's noise and therefore gives poor performance as evident from a high reprojection error for all sensor pairs (Table \ref{table: line_rep_err_1}). To prove our hypothesis we re-calibrate our sensors using \textbf{MSG-Cal} but with the Ouster LIDAR removed and the results are presented in Table \ref{table: line_rep_err_2} and Figure \ref{fig: vlp_left_withAndwithout_ouster}.
\begin{table}[!ht]
\setlength{\tabcolsep}{2.25pt}
\centering
\begin{tabular}{| c | c | c | c |} 
\hline
  \textbf{3D-LIDAR Camera Pair} & \multicolumn{3}{|c|}{MLRE}\\ 
 \hline
  & \textbf{PPC-Cal} & \textbf{PBPC-Cal} & \textbf{MSG-Cal} \\ [0.5ex] 
 \hline
 \texttt{VLP-32 $\leftrightarrow$ Stereo Left} & 2.57316 & \textcolor{blue}{1.94707} & \textcolor{red}{3.25161} \\ \hline
 \texttt{VLP-32 $\leftrightarrow$ Stereo Right} & \textcolor{red}{2.94664} & \textcolor{blue}{1.88719} & 2.77772 \\ \hline
 \texttt{VLP-32 $\leftrightarrow$ Basler} & 2.21383 & \textcolor{blue}{1.9423} & \textcolor{red}{2.5893} \\ \hline
\end{tabular}
\caption{MLRE (in pixels) for various 3D-LIDAR Camera Pairs with \textbf{PPC-Cal}, \textbf{PBPC-Cal} and \textbf{MSG-Cal} without Ouster OS1 64 LIDAR}
\label{table: line_rep_err_2}
\end{table}

From Figure \ref{fig: vlp_left_withAndwithout_ouster} we can conclude that \textbf{MSG-Cal} shows significant improvement when used in the absence of Ouster LIDAR and Table \ref{table: line_rep_err_2} conveys that without the Ouster in the graph optimization framework, the results of \textbf{MSG-Cal} are similar to those of \textbf{PPC-Cal} which makes sense because the pair-wise calibration in \textbf{MSG-Cal} uses similar constraints as \textbf{PPC-Cal}. Irrespective of Ouster LIDAR's presence or absence, \textbf{PBPC-Cal} performs the best. 

\section{Discussion}
In this work we compared three LIDAR Camera extrinsic calibration algorithms \emph{viz.} \textbf{PPC-Cal} (\cite{unnikrishnan2005}, \cite{Huang2009} \& \cite{PANDEY2010336}), \textbf{PBPC-Cal} \cite{subodhIV2020} and \textbf{MSG-Cal} \cite{Owens_msg-cal:multi-sensor}. We presented the mathematical framework behind the working of all these methods and extensively evaluated them on a multi-sensor platform comprising 3 distinct cameras and 2 LIDARs. We concluded that \textbf{PPC-Cal} \& \textbf{PBPC-Cal} were robust to random initialization for all trials while \textbf{MSG-Cal} diverged in a few trials (Figure \ref{fig: randomInitialization}). Nevertheless, barring a few cases in \textbf{MSG-Cal}, all the three frameworks can be initialized with any intial condition and will still converge. We showed that the \textbf{PPC-Cal} which uses only \textit{point to plane} constraint shows deterioration in performance when a noisy sensor is used (Table \ref{table: line_rep_err_1}). The use of additional \textit{point to back-projected} plane constraint in \textbf{PBPC-Cal} helps reduce the effect of noisy sensor by introducing more geometrical constraints to the non-linear cost function. We also showed that the global graph based optimization method \textbf{MSG-Cal}, which uses a variant of the \textit{point to plane} constraint has all final pair wise calibrations (as evident from high MLRE from Table \ref{table: line_rep_err_1}) affected in the presence of a noisy sensor but gives comparable performance to \textbf{PPC-Cal} when the noisy sensor is removed (Table \ref{table: line_rep_err_2}). \textbf{PBPC-Cal} exhibits similar performance both with and without the noisy sensor and performs better than both \textbf{PPC-Cal} and \textbf{MSG-Cal} under all circumstances (Tables \ref{table: line_rep_err_1} \& \ref{table: line_rep_err_2}, Figure \ref{fig: comparisonABC}). If we do not have a noisy sensor and need a quick calibration result then using \textbf{PPC-Cal} is a good option but if we have multiple sensors (with low noise) then \textbf{MSG-Cal} should be the algorithm of choice, as collecting data for both these methods is easier. Instead, if we have noisy sensors then \textbf{PBPC-Cal} should be used. In the future we want to use \textbf{PBPC-Cal} in the pair-wise calibration step of \textbf{MSG-Cal}, thus bringing the benefits of robust pair-wise calibration and joint global optimization together.

\section{Acknowledgements}
The authors would like to thank Peng Jiang for his invaluable help with data collection.
\bibliographystyle{IEEEtran}
\bibliography{bibexpendable}

\begin{thebibliography}{10}
\providecommand{\url}[1]{#1}
\csname url@samestyle\endcsname
\providecommand{\newblock}{\relax}
\providecommand{\bibinfo}[2]{#2}
\providecommand{\BIBentrySTDinterwordspacing}{\spaceskip=0pt\relax}
\providecommand{\BIBentryALTinterwordstretchfactor}{4}
\providecommand{\BIBentryALTinterwordspacing}{\spaceskip=\fontdimen2\font plus
\BIBentryALTinterwordstretchfactor\fontdimen3\font minus
  \fontdimen4\font\relax}
\providecommand{\BIBforeignlanguage}[2]{{%
\expandafter\ifx\csname l@#1\endcsname\relax
\typeout{** WARNING: IEEEtran.bst: No hyphenation pattern has been}%
\typeout{** loaded for the language `#1'. Using the pattern for}%
\typeout{** the default language instead.}%
\else
\language=\csname l@#1\endcsname
\fi
#2}}
\providecommand{\BIBdecl}{\relax}
\BIBdecl

\bibitem{unnikrishnan2005}
R.~Unnikrishnan and M.~Hebert, ``Fast extrinsic calibration of a
  laserrangefinder to a camera,'' 07 2005.

\bibitem{Huang2009}
L.~{Huang} and M.~{Barth}, ``A novel multi-planar lidar and computer vision
  calibration procedure using 2d patterns for automated navigation,'' in
  \emph{2009 IEEE Intelligent Vehicles Symposium}, June 2009, pp. 117--122.

\bibitem{PANDEY2010336}
\BIBentryALTinterwordspacing
G.~Pandey, J.~McBride, S.~Savarese, and R.~Eustice, ``Extrinsic calibration of
  a 3d laser scanner and an omnidirectional camera,'' \emph{IFAC Proceedings
  Volumes}, vol.~43, no.~16, pp. 336 -- 341, 2010, 7th IFAC Symposium on
  Intelligent Autonomous Vehicles. [Online]. Available:
  \url{http://www.sciencedirect.com/science/article/pii/S1474667016350790}
\BIBentrySTDinterwordspacing

\bibitem{Zhou201206}
L.~Zhou and Z.~Deng, ``Extrinsic calibration of a camera and a lidar based on
  decoupling the rotation from the translation,'' 06 2012, pp. 642--648.

\bibitem{Zhou201810}
L.~Zhou, Z.~Li, and M.~Kaess, ``Automatic extrinsic calibration of a camera and
  a 3d lidar using line and plane correspondences,'' 10 2018, pp. 5562--5569.

\bibitem{PANDEY201409}
G.~Pandey, J.~McBride, S.~Savarese, and R.~Eustice, ``Automatic extrinsic
  calibration of vision and lidar by maximizing mutual information,''
  \emph{Journal of Field Robotics}, vol.~32, 09 2014.

\bibitem{Levinson201306}
J.~Levinson and S.~Thrun, ``Automatic online calibration of cameras and
  lasers,'' 06 2013.

\bibitem{scaramuzza}
D.~{Scaramuzza}, A.~{Harati}, and R.~{Siegwart}, ``Extrinsic self calibration
  of a camera and a 3d laser range finder from natural scenes,'' in \emph{2007
  IEEE/RSJ International Conference on Intelligent Robots and Systems}, Oct
  2007, pp. 4164--4169.

\bibitem{Taylor201605}
Z.~{Taylor} and J.~{Nieto}, ``Motion-based calibration of multimodal sensor
  extrinsics and timing offset estimation,'' \emph{IEEE Transactions on
  Robotics}, vol.~32, no.~5, pp. 1215--1229, Oct 2016.

\bibitem{Strauss1995}
S.~{Wasielewski} and O.~{Strauss}, ``Calibration of a multi-sensor system laser
  rangefinder/camera,'' in \emph{Proceedings of the Intelligent Vehicles '95.
  Symposium}, Sep. 1995, pp. 472--477.

\bibitem{Zhang1389752}
{Qilong Zhang} and R.~{Pless}, ``Extrinsic calibration of a camera and laser
  range finder (improves camera calibration),'' in \emph{2004 IEEE/RSJ
  International Conference on Intelligent Robots and Systems (IROS) (IEEE Cat.
  No.04CH37566)}, vol.~3, Sep. 2004, pp. 2301--2306 vol.3.

\bibitem{Ruben7139700}
R.~{Gomez-Ojeda}, J.~{Briales}, E.~{Fernandez-Moral}, and
  J.~{Gonzalez-Jimenez}, ``Extrinsic calibration of a 2d laser-rangefinder and
  a camera based on scene corners,'' in \emph{2015 IEEE International
  Conference on Robotics and Automation (ICRA)}, May 2015, pp. 3611--3616.

\bibitem{Naroditsky2011}
O.~{Naroditsky}, A.~{Patterson}, and K.~{Daniilidis}, ``Automatic alignment of
  a camera with a line scan lidar system,'' in \emph{2011 IEEE International
  Conference on Robotics and Automation}, May 2011, pp. 3429--3434.

\bibitem{Bonnifait4648067}
S.~A. {Rodriguez F.}, V.~{Fremont}, and P.~{Bonnifait}, ``Extrinsic calibration
  between a multi-layer lidar and a camera,'' in \emph{2008 IEEE International
  Conference on Multisensor Fusion and Integration for Intelligent Systems},
  Aug 2008, pp. 214--219.

\bibitem{butvelodyne}
M.~Velas, M.~Spanel, Z.~Materna, and A.~Herout, ``Calibration of rgb camera
  with velodyne lidar.''

\bibitem{Geiger}
A.~Geiger, F.~Moosmann, O.~Car, and B.~Schuster, ``Automatic camera and range
  sensor calibration using a single shot,'' \emph{Proceedings - IEEE
  International Conference on Robotics and Automation}, pp. 3936--3943, 05
  2012.

\bibitem{subodhIV2020}
S.~Mishra, G.~Pandey, and S.~Saripalli, ``Extrinsic calibration of a 3d-lidar
  and a camera,'' \emph{Submitted to IV2020}, 02 2020.

\bibitem{jointCalNg}
Q.~V. {Le} and A.~Y. {Ng}, ``Joint calibration of multiple sensors,'' in
  \emph{2009 IEEE/RSJ International Conference on Intelligent Robots and
  Systems}, Oct 2009, pp. 3651--3658.

\bibitem{Owens_msg-cal:multi-sensor}
J.~L. Owens, P.~R. Osteen, E.~Corporation, and K.~Daniilidis, ``Msg-cal:
  Multi-sensor graph-based calibration.''

\bibitem{articleAruco1}
S.~Garrido-Jurado, R.~Mu\~{n}oz Salinas, F.~Madrid-Cuevas, and
  R.~Medina-Carnicer, ``Generation of fiducial marker dictionaries using mixed
  integer linear programming,'' \emph{Pattern Recognition}, vol.~51, 10 2015.

\bibitem{olson2011tags}
E.~Olson, ``{AprilTag}: A robust and flexible visual fiducial system,'' in
  \emph{Proceedings of the {IEEE} International Conference on Robotics and
  Automation ({ICRA})}.\hskip 1em plus 0.5em minus 0.4em\relax IEEE, May 2011,
  pp. 3400--3407.

\bibitem{spherical2018}
J.~{Kümmerle}, T.~{Kühner}, and M.~{Lauer}, ``Automatic calibration of
  multiple cameras and depth sensors with a spherical target,'' in \emph{2018
  IEEE/RSJ International Conference on Intelligent Robots and Systems (IROS)},
  Oct 2018, pp. 1--8.

\bibitem{zhang-camera-calib}
\BIBentryALTinterwordspacing
Z.~Zhang, ``A flexible new technique for camera calibration,'' \emph{IEEE
  Trans. Pattern Anal. Mach. Intell.}, vol.~22, no.~11, p. 1330–1334, Nov.
  2000. [Online]. Available: \url{https://doi.org/10.1109/34.888718}
\BIBentrySTDinterwordspacing

\bibitem{Hartley2003MVG861369}
R.~Hartley and A.~Zisserman, \emph{Multiple View Geometry in Computer Vision},
  2nd~ed.\hskip 1em plus 0.5em minus 0.4em\relax New York, NY, USA: Cambridge
  University Press, 2003.

\bibitem{RANSAC}
\BIBentryALTinterwordspacing
M.~A. Fischler and R.~C. Bolles, ``Random sample consensus: A paradigm for
  model fitting with applications to image analysis and automated
  cartography,'' \emph{Commun. ACM}, vol.~24, no.~6, pp. 381--395, Jun. 1981.
  [Online]. Available: \url{http://doi.acm.org/10.1145/358669.358692}
\BIBentrySTDinterwordspacing

\bibitem{PCL}
R.~B. {Rusu} and S.~{Cousins}, ``3d is here: Point cloud library (pcl),'' in
  \emph{2011 IEEE International Conference on Robotics and Automation}, May
  2011, pp. 1--4.

\bibitem{opencvlibrary}
G.~Bradski, ``{The OpenCV Library},'' \emph{Dr. Dobb's Journal of Software
  Tools}, 2000.

\bibitem{ceres-solver}
S.~Agarwal, K.~Mierle, and Others, ``Ceres solver,''
  \url{http://ceres-solver.org}.

\bibitem{LSD4731268}
R.~{Grompone von Gioi}, J.~{Jakubowicz}, J.~{Morel}, and G.~{Randall}, ``Lsd: A
  fast line segment detector with a false detection control,'' \emph{IEEE
  Transactions on Pattern Analysis and Machine Intelligence}, vol.~32, no.~4,
  pp. 722--732, April 2010.

\bibitem{PribylZC16a}
\BIBentryALTinterwordspacing
B.~Pribyl, P.~Zemc{\'{\i}}k, and M.~Cad{\'{\i}}k, ``Pose estimation from line
  correspondences using direct linear transformation,'' \emph{CoRR}, vol.
  abs/1608.06891, 2016. [Online]. Available:
  \url{http://arxiv.org/abs/1608.06891}
\BIBentrySTDinterwordspacing

\bibitem{g2o}
R.~Kümmerle, G.~Grisetti, H.~Strasdat, K.~Konolige, and W.~Burgard, ``G2o: A
  general framework for graph optimization,'' 06 2011, pp. 3607 -- 3613.

\end{thebibliography}

\end{document}